\pdfoutput=1

\documentclass[11pt]{article}

\usepackage[]{ACL2023}

\usepackage{times}
\usepackage{latexsym}
\usepackage{subcaption}
\usepackage{caption}
\usepackage{lipsum}

\usepackage[T1]{fontenc}

\usepackage[utf8]{inputenc}

\usepackage{microtype}

\usepackage{inconsolata}
\pdfoutput=1
\usepackage{graphicx}
\usepackage{hyperref}
\usepackage{multicol}
\usepackage{multirow}
\usepackage{booktabs}

\title{Another Dead End for Morphological Tags? Perturbed Inputs and Parsing}

\author{Alberto Muñoz-Ortiz and David Vilares\\
  Universidade da Coruña, CITIC \\
  Departamento de Ciencias de la Computación y Tecnologías de la Información\\
  Campus de Elviña s/n, 15071\\
  A Coruña, Spain \\
  \texttt{\{alberto.munoz.ortiz, david.vilares\}@udc.es} \\}

\begin{document}
\maketitle
\begin{abstract}

The usefulness of part-of-speech tags for parsing has been heavily questioned due to the success of word-contextualized parsers. Yet, most studies are limited to coarse-grained tags and high quality written content; while we know little about their influence when it comes to models in production that face lexical errors. We expand these setups and design an adversarial attack to verify if the use of morphological information by parsers: (i) contributes to error propagation or (ii) if on the other hand it can play a role to correct mistakes that word-only neural parsers make. The results on 14 diverse UD treebanks show that under such attacks, for transition- and graph-based models their use contributes to degrade the performance even faster, while for the (lower-performing) sequence labeling parsers they are helpful. We also show that if morphological tags were utopically robust against lexical perturbations, they would be able to correct parsing mistakes.

\end{abstract}

\section{Introduction}

The use of morphological tags was a core component of dependency parsers to improve performance \cite{ballesteros-nivre-2012-maltoptimizer-system}. With the rise of neural models, feeding explicit morphological information is a practice that has greatly vanished, with (often) the exception of part-of-speech (PoS) tags. In this line, \citet{ballesteros-etal-2015-improved} already found that character-based word vectors helped improving performance over purely word-level models, specially for rich-resource languages, for which the use of morphological information is more relevant \cite{dehouck-denis-2018-framework}. Related, \citet{dozat-etal-2017-stanfords} showed that predicted PoS tags still improved the performance of their graph-based parser, even when used together with character-based representations. \citet{smith-etal-2018-investigation} and \citet{de-lhoneux-etal-2017-raw} studied the impact that ignoring PoS tag vectors had on the performance of a biLSTM transition-based parser \cite{kiperwasser-goldberg-2016-simple}. They conclude that when considering PoS tags, word-level, and character-level embedddings, any two of those vectors are enough to maximize a parser performance, i.e., PoS tag vectors can be excluded when using \emph{both} word-level and character-level vectors. \citet{zhou2020pos} showed the utility of PoS tags when learned jointly with parsing. Recently, \citet{anderson-gomez-rodriguez-2021-taggers} and \citet{anderson-etal-2021-falta} have explored the differences between using gold and predicted PoS tags, showing that the former are helpful to improve the results, while the latter are often not, with the exception of low-resource languages, where they obtain small but consistent improvements. Furthermore, \citet{munoz-ortiz-etal-2022-parsing} showed that the efficacy of PoS tags in the context of sequence labeling parsing is greatly influenced by the chosen linearization method.

However, most of such work has focused on: (i) studying the effect of the universal PoS tags \cite{ud2.9}, and (ii) its impact on non-perturbed inputs. Yet, NLP models are very sensible and brittle against small attacks, and simple perturbations like misspellings can greatly reduce performance \cite{ebrahimi-etal-2018-hotflip,alzantot-etal-2018-generating}. This has been shown for tasks such as named-entity recognition, question answering, semantic similarity, and sentiment analysis \cite{moradi-samwald-2021-evaluating}. In parallel, defensive strategies have been tested to improve the robustness of NLP systems, e.g., placing a word recognition module before downstream classifiers \cite{pruthi-etal-2019-combating}, or using spelling checks and adversarial training \cite{li-etal-2019-textbugger}. Yet, as far as we know, no related work has been done on testing perturbed inputs for parsing and the effect, positive or negative, that using morphological information as explicit signals during inference might have in guiding the parsers.\footnote{The code related to this work is available at \url{https://github.com/amunozo/parsing_perturbations}.}

\section{Adversarial framework}
Perturbed inputs occur for several reasons, such as for instance on-purpose adversarial attacks \cite{LiangDeep2018} or, more likely, unintended mistakes made by human writers. In any case, they have an undesirable effect on NLP tools, including parsers. Our goal is to test if under such adversarial setups, coarse- and fine-grained morphological tags: (i) could help obtaining more robust and better results in comparison to word-only parsers (going against the current trend of removing any explicit linguistic input from parsers); or (ii) if on the contrary they contribute to degrade parsing performance.

Below, we describe both how we generate (i, \S \ref{ssec:perturbation}) linguistically-inspired attacks at character-level, and (ii, \S \ref{ssec:parsing_models}) the tested parsers.

\subsection{Perturbed inputs}\label{ssec:perturbation}

To perturb our inputs, we use a combination of four adversarial misspellings, inspired by \citet{pruthi-etal-2019-combating} who designed their method relying on previous psycholinguistic studies \cite{davis-2003-psycholinguistic,rawlinson1976significance}. In particular, we consider to: (i) drop one character, (ii) swap two contiguous characters, (iii) add one character, and (iv) replace a character with an adjacent character in a QWERTY keyboard. These changes will probably transform most words into out-of-vocabulary terms, although some perturbations could generate valid tokens (likely occurring in an invalid context). 
We only apply perturbations to a fraction of the content words of a sentence\footnote{Those which universal PoS tags is \texttt{ADJ}, \texttt{ADV}, \texttt{INTJ}, \texttt{PROPN}, \texttt{NOUN} or \texttt{VERB}.} (details in \S \ref{sec:experiments}), as function words tend to be shorter and a 
perturbation could make them unrecognizable, which is not our aim.

Finally, we only allow a word to suffer a single attack. Since we will be evaluating on a multilingual setup, we considered language-specific keyboards to generate the perturbations. We restrict our analysis to languages that use the Latin alphabet, but our adversarial attack would be, in principle, applicable to any alphabetic script.

\subsection{Parsing models}\label{ssec:parsing_models}

Since we want a thorough picture of the impact of using morphological information on parsers, we include three models from different paradigms:

\begin{enumerate}
\item A left-to-right transition-based parser with pointer networks \cite{fernandez-gonzalez-gomez-rodriguez-2019-left}. It uses biLSTMs \cite{hochreiter1997long} to contextualize the words, and the outputs are then fed to a pointer network \cite{vinyals2015pointer}, which keeps a stack and, in a left-to-right fashion, decides for each token its head.
\item A biaffine graph-based parser \cite{dozat-etal-2017-stanfords}. This model also uses biLSTMs to first contextualize the input sentence. Differently from \citeauthor{fernandez-gonzalez-gomez-rodriguez-2019-left}, the tree is predicted through a biaffine attention module, and to ensure well-formed trees it uses either the \citet{eisner-1996-three} or \citet{chu1965shortest,edmonds1968optimum} algorithms.\footnote{This is true for the \texttt{supar} implementation that we use, although \citeauthor{dozat-etal-2017-stanfords} relied on heuristics.}

\item A sequence labeling parser
\cite{strzyz-etal-2020-bracketing} that uses a 2-planar bracketing encoding to linearize the trees.
Like the two other parsers, it uses  biLSTMs to contextualize sentences, but it does not use any mechanism on top of their outputs (such as biaffine attention or a decoder module) to predict the tree (which is rebuilt from a sequence of labels). 
\end{enumerate}

Particularly, we use this third model to: (i) estimate how sensitive raw biLSTMs are to attacks, (ii) compare their behavior against the transition- and graph-based models and the extra mechanisms that they incorporate, (iii) and verify if such mechanisms play a role against perturbed inputs.

\paragraph{Inputs} We concatenate a word vector, a second word vector computed at character level, and (optionally) a morphological vector. This is the preferred input setup of previous work on PoS tagging plus its utility for neural UD parsing \cite{de-lhoneux-etal-2017-raw,anderson-gomez-rodriguez-2021-taggers}.\footnote{Some authors \cite{zhou2020pos} exploit PoS tags for parsing in a multi-task learning setup instead, but the differences in the experiments are small ($\sim$0.3 points) and they are limited to English and Chinese on non-UD treebanks.} Note that character-level vectors should be robust against our attacks, but it is known that in practice they are fragile \cite{pruthi-etal-2019-combating}.
In this respect, our models use techniques to strengthen their behaviour against word variation, by
using character-level dropout.
This way, we inject noise during training and give all our models a lexical-level defensive mechanism to deal with misspellings. We kept this feature to keep the setup realistic, as character-level dropout is implemented by default in most of modern parsers, and ensure stronger baselines.

\paragraph{Training and hyperparameters} We use non-perturbed training and development sets,\footnote{For the models that use morphological information we went for gold tags for training. 
The potential advantages of training with predicted PoS tags vanish here, as the error distribution for PoS tags would be different for non-perturbed (during training) \emph{versus} perturbed inputs (during testing).
} since our aim is to see how parsers trained in a standard way (and that may use explicit morphological features) behave in production under adversarial attacks. 
Alternatively, we could design additional techniques to protect the parsers against such perturbations, but this is out of the scope of this paper (and for standard defensive strategies, we already have character-level dropout). For all parsers, we use the default configuration specified in the corresponding repositories. We use 2 GeForce RTX 3090 for training the models for around 120 hours.

\paragraph{Morphological tags}
To predict them, we use a sequence labeling model with the same architecture than 
the one used for the sequence labeling parser. We use as input a concatenation of a word embedding and a character-level LSTM vector.

\section{Experiments}\label{sec:experiments}

We now describe our experimental setup:

\paragraph{Data} We selected 14 UD treebanks \cite{ud2.9} that use the Latin alphabet and are annotated with universal PoS tags (UPOS), language-specific PoS tags (XPOS), and morphological feats (FEATS). It is a diverse sample that considers different language families and amounts of data, whose details are shown in Table \ref{tab:tb_info}.
For the pre-trained word vectors, we rely on \citet{bojanowski2017enriching}.\footnote{We exclude experiments with BERT-based models for a few reasons: (i) to be homogeneous with previous setups (e.g. \citet{smith-etal-2018-investigation}, \citet{anderson-etal-2021-falta}), (ii) because the chosen parsers already obtain competitive results without the need of these models, and (iii) for a better understanding of the results, since it is hard to interpret the performances of individual languages while not extracting conclusions biased on the language model used, instead of the parsing architecture.
} Also, note that we only perturb the test inputs.
Thus, when the input is highly perturbed, the model will mostly depend on the character representations, and if used, the morphological tags fed to it.

\begin{table}[htpb]
\scriptsize
    \centering
    \begin{tabular}{lrllrrr}
    \hline
        Treebank & \# Sent. & Family & \#UPOS & \#XPOS &  \#FEATS \\
        \hline
        \tiny{Afrikaans\textsubscript{AfriBooms}} & 1\,315 & Germanic (IE) & 16 & 95 & 55 \\
        Basque\textsubscript{BDT} & 5\,396 & Basque & 16 & - & 573 \\
        English\textsubscript{EWT} & 12\,543 & Germanic (IE) & 18 & 51 & 153 \\
        Finnish\textsubscript{TDT} & 12\,217 & Uralic & 16 & 14 & 1\,786\\
        German\textsubscript{GSD} & 13\,814 & Germanic (IE) & 17 & 52 & 458 \\
        Hungarian\textsubscript{Szeged} & 449 & Uralic & 16 & - & 384\\
        Indonesian\textsubscript{GSD} & 4\,477& Austronesian & 18 & 45 & 48\\
        Irish\textsubscript{IDT} & 4\,005 & Celtic (IE) & 17 & 72 & 653 \\
        Lithuanian\textsubscript{HSE} & 153  & Baltic (IE) & 16 & 30 & 215 \\
        Maltese\textsubscript{MUDT} & 1\,123 & Afro-Asiatic & 17 & 47 & - \\
        Polish\textsubscript{LFG} & 13\,774  & Slavic (IE) & 15 & 623 & 1\,037\\
        Spanish\textsubscript{AnCora} & 14\,305 & Latin (IE) & 18 & 318 & 243\\
        Swedish\textsubscript{LinES} & 3\,176 & Germanic (IE) & 17 & 214 & 171\\
        Turkish\textsubscript{Penn} & 14\,851 & Turkic & 15 & - & 490\\ \hline
    \end{tabular}
    \caption{Relevant information for the treebanks used.}
    \label{tab:tb_info}
\end{table}

\paragraph{Generating perturbed treebanks}
For each test set, we create several versions with increasing percentages of perturbed content words (from 0\% to 100\%, with steps of 10 percent points) to monitor how the magnitude of the attacks affects the results. For each targeted word, one of the four proposed perturbations is applied randomly. To control for randomness, each model is tested against 10 perturbed test sets with the same level of perturbation. To check that the scores were similar across runs, we computed the average scores and the standard deviation (most of them exhibiting low values).

\paragraph{Setup}
For each parser we trained four models: 
a word-only (\texttt{word}) baseline where the input is just the concatenation of a pre-trained word vector and a character-level vector, and \emph{three} extra models that use universal PoS tags (\texttt{word+UPOS}), language-specific PoS tags (\texttt{word+XPOS}), or feats (\texttt{word+FEATS}). 
For parsing evaluation, we use 
labeled attachment scores (LAS). For the taggers, we report accuracy. 
We evaluate the models on two setups regarding the prediction of morphological tags: (i) tags predicted on the same perturbed inputs as the dependency tree, and (ii) tags predicted on non-perturbed inputs. Specifically, the aim of setup ii is to simulate the impact of using a tagger that is very robust against lexical perturbations.

\begin{table*}[!]
    \centering
    \scriptsize
    \begin{tabular}{c|llll|llll|llll||lll}
    \hline
    \multirow{2}{*}{\% Perturbed} & \multicolumn{4}{c|}{Transition-based} & \multicolumn{4}{c|}{Graph-based} & \multicolumn{4}{c||}{Sequence labeling}  & \multicolumn{3}{c}{Tagger accuracy}\\
    & \texttt{word} & \texttt{UPOS} & \texttt{XPOS} & \texttt{FEATS} & \texttt{word} & \texttt{UPOS} & \texttt{XPOS} & \texttt{FEATS} & \texttt{word} & \texttt{UPOS} & \texttt{XPOS} & \texttt{FEATS} & UPOS & XPOS & FEATS \\
        \hline
        0  & 75.66 & 74.93 & 76.28 & 74.84 & 79.35 & 77.44 & 78.38 & 77.28 & 68.29 & 68.98 & 70.96 & 66.79 & 89.76 & 87.80 & 83.38\\
        10  & 74.93 & 73.68  & 75.07 & 73.53 & 78.59 & 75.69 & 76.77 & 75.49 & 66.71 & 67.31 & 69.34 & 64.97 & 88.56 & 86.17 & 81.68\\
        20  & 74.11 & 72.45 & 73.92 & 72.13 & 77.81 & 73.93 & 75320 & 73.73 & 65.18 & 65.61 & 67.76 & 63.16 & 87.38 & 84.59 & 79.94\\
        30  & 73.33 & 71.19 & 72.66 & 70.74 & 76.99 & 72.22 & 73.56 & 71.92 & 63.62 & 63.96 & 66.17 & 61.37 & 86.17 & 82.91 & 78.22 \\
        40  & 72.52 & 69.86 & 71.45 & 69.33 & 76.10 & 70.36 & 71.88 & 70.06 & 62.09 & 62.24 & 64.59 & 59.55 & 84.93 & 81.30 & 76.50 \\
        50  & 71.66 & 68.58 & 70.13 & 67.93 & 75.27 & 68.63 & 70.14 & 68.09 & 60.52 & 60.50 & 62.94 & 57.81 & 83.71 & 79.61 & 74.68\\
        60  & 70.78 & 67.26 & 68.75 & 66.46 & 74.37 & 66.72 & 68.37 & 66.09 & 58.94 & 58.91 & 61.36 & 56.10 & 82.48 & 77.90 & 72.92\\
        70  & 69.87 & 65.88 & 67.40 & 64.92 & 73.49 & 64.96 & 66.64 & 66.06 & 57.44 & 57.24 & 59.77 & 54.36 & 81.19 & 76.13 & 71.13\\
        80  & 68.96 & 64.50 & 66.03 & 63.46 & 72.48 & 63.05 & 64.80 & 62.27 & 55.90 & 55.61 & 58.17 & 52.65 & 79.93 & 74.42 & 69.37\\
        90  & 67.99 & 63.12 & 64.61 & 61.90 & 71.57 & 61.12 & 62.97 & 60.16 & 54.42 & 53.95 & 56.54 & 50.96 & 78.62 & 72.64 & 67.56\\
        100 & 67.04 & 61.74 & 63.16 & 60.34 & 70.59 & 59.23 & 61.14 & 58.13 & 52.92 & 52.30 & 54.97 & 49.23 & 77.30 & 70.85 & 65.74\\
        \hline
    \end{tabular}
    \caption{On the left, average LAS scores for all treebanks and degrees of perturbation for the \texttt{word}, \texttt{word+UPOS}, \texttt{word+XPOS}, and \texttt{word+FEATS} models \emph{using morphological tags predicted on perturbed input}. On the right, the average scores for the taggers used.
    }
    \label{tab:parsers_las}
\end{table*}

\begin{table*}[!]
    \centering
    \scriptsize
    \begin{tabular}{c|llll|llll|llll}
    \hline
    \multirow{2}{*}{\% Perturbed} & \multicolumn{4}{c|}{Transition-based} & \multicolumn{4}{c|}{Graph-based} & \multicolumn{4}{c}{Sequence labeling} \\
    & \texttt{word} & \texttt{UPOS} & \texttt{XPOS} & \texttt{FEATS} & \texttt{word} & \texttt{UPOS} & \texttt{XPOS} & \texttt{FEATS} & \texttt{word} & \texttt{UPOS} & \texttt{XPOS} &  \\
        \hline
        0  & 75.66 & 74.93 & 76.28 & 74.84 & 79.35 & 77.44 & 78.38 & 77.28 & 68.29 & 68.98 & 70.96 & 66.79 \\
        10  & 74.93 & 74.64 & 76.05 & 74.55 & 78.59 & 76.91 & 78.01 & 76.78 & 66.71 & 68.60 & 70.53 & 66.19 \\
        20  & 74.11 & 74.36 & 75.82 & 74.23 & 77.81 & 76.46 & 77.58 & 73.62 & 65.18 & 68.19 & 70.08 & 65.62 \\
        30  & 73.33 & 74.02 & 75.60 & 73.94 & 76.99 & 75.88 & 77.20 & 75.82 & 63.62 & 67.76 & 69.62 & 64.99 \\
        40  & 72.52 & 73.71 & 75.36 & 73.66 & 76.10 & 75.44 & 76.78 & 75.27 & 62.09 & 67.34 & 69.13 & 64.46 \\
        50  & 71.66 & 73.41 & 75.17 & 73.35 & 75.27 & 74.94 & 76.42 & 74.80 & 60.52 & 66.88 & 68.66 & 63.79 \\
        60  & 70.78 & 73.06 & 74.87 & 73.04 & 74.37 & 74.46 & 76.02 & 74.25 & 58.94 & 66.40 & 68.19 & 63.18 \\
        70  & 69.87 & 72.74 & 74.64 & 72.70 & 73.49 & 73.99 & 75.53 & 73.76 & 57.44 & 65.95 & 67.72 & 62.56 \\
        80  & 69.86 & 72.39 & 74.40 & 72.37 & 72.48 & 73.46 & 75.13 & 73.26 & 55.90 & 65.45 & 67.23 & 61.92 \\
        90  & 67.99 & 72.08 & 74.13 & 72.10 & 71.57 & 72.92 & 74.46 & 72.73 & 54.42 & 64.93 & 66.75 & 61.27 \\
        100 & 67.04 & 71.73 & 73.93 & 71.74 & 70.59 & 72.45 & 74.35 & 72.15 & 52.92 & 64.41 & 66.27 & 60.63 \\
        \hline
    \end{tabular}
    \caption{Average LAS scores for all treebanks and degrees of perturbation for the \texttt{word}, \texttt{word+UPOS}, \texttt{word+XPOS}, and \texttt{word+FEATS} models \emph{using morphological tags predicted on non-perturbed input}.
    }
    \label{tab:las_unperturbed}
\end{table*}

\subsection{Results}
\begin{figure}
\centering
\begin{subfigure}{0.23\textwidth}
  \centering
  \includegraphics[width=\textwidth]{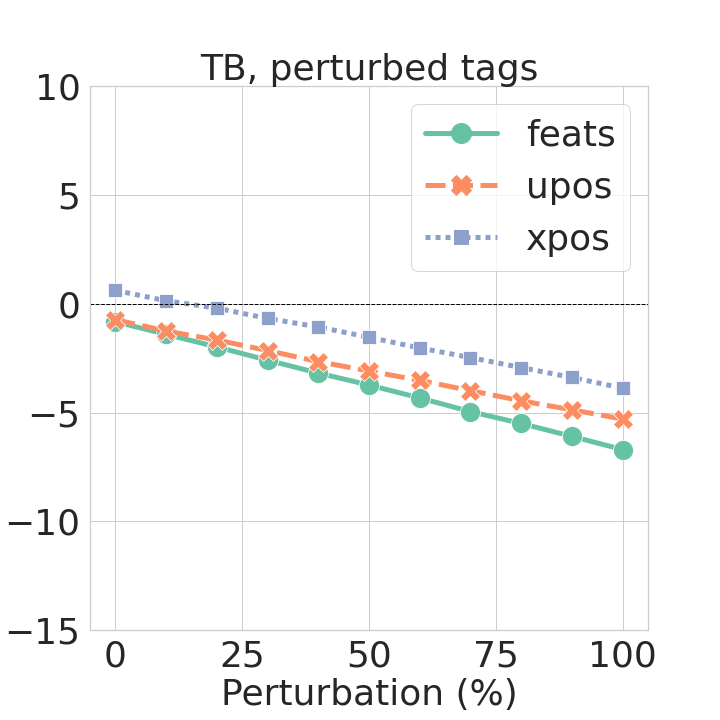}
  \label{subfig:tb_results_a}
  \caption{Perturbed}
\end{subfigure}
\hfill
\begin{subfigure}{0.23\textwidth}
  \centering
  \includegraphics[width=\textwidth]{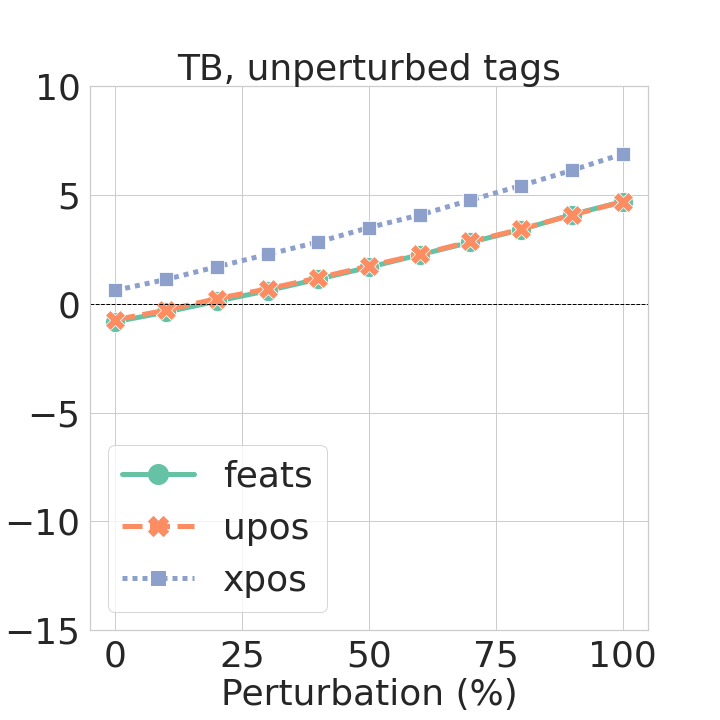}
  \label{fig:tb_results_b}
  \caption{Non-perturbed}
\end{subfigure}
    \caption{Average $\Delta$LAS across all treebanks for the transition-based models \texttt{word+upos}, \texttt{word+xpos}, and \texttt{word+feats} vs \texttt{word}, using morphological tags predicted on perturbed and non-perturbed inputs.}
    \label{fig:tb_results}
\end{figure}

Tables \ref{tab:parsers_las} and \ref{tab:las_unperturbed} show the average LAS results across all treebanks and models for tags predicted on perturbed and non-perturbed inputs, respectively.
Figures \ref{fig:tb_results}, \ref{fig:gb_results}, and \ref{fig:sl_results} display the mean LAS difference between the \texttt{word} and the other model configurations, using tags predicted on both perturbed and non-perturbed inputs for each parser.

\subsubsection{Results using morphological tags predicted on perturbed inputs}
Figure \ref{fig:tb_results}.a shows the score differences for the transition-based parsers.
The average difference between the baseline and all the models using morphological tags becomes more negative as the percentage of perturbed words increases. 
Such difference is only positive for \texttt{word+XPOS} when none or a few percentage of words are perturbed. All
morphological tags show a similar tendency, 
\texttt{word+FEATS} degrading the performance the most, followed by 
the `coarse-grained' \texttt{word+UPOS}.

\begin{figure}
\centering
\begin{subfigure}{0.23\textwidth}
  \centering
  \includegraphics[width=\textwidth]{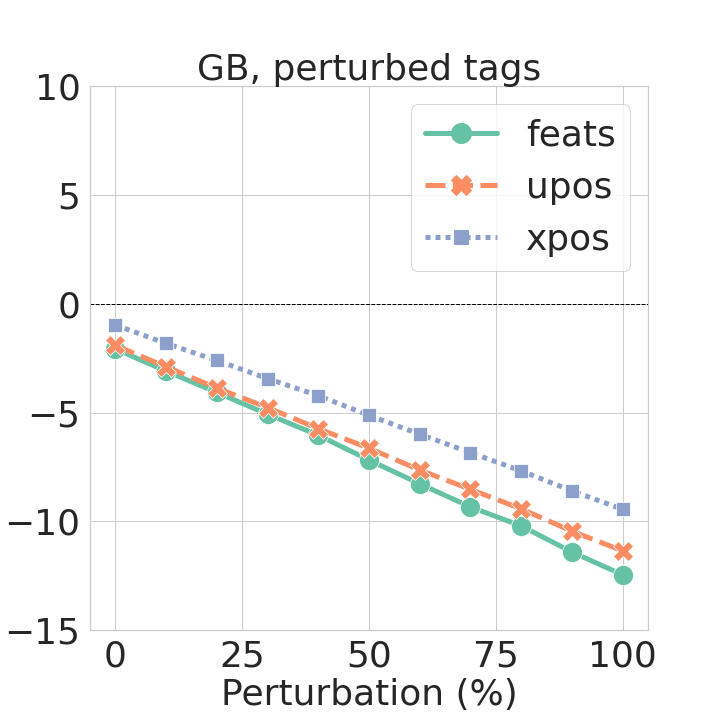}
  \label{fig:image1}
  \caption{Perturbed}
\end{subfigure}
\hfill
\begin{subfigure}{0.23\textwidth}
  \centering
  \includegraphics[width=\textwidth]{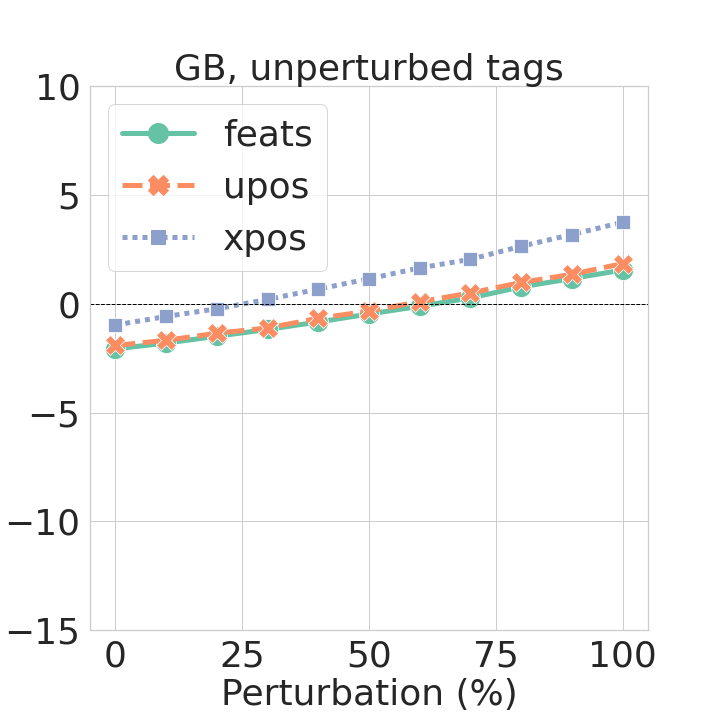}
  \label{fig:image2}
  \caption{Non-perturbed}
\end{subfigure}
    \caption{Average $\Delta$LAS across all treebanks for the graph-based models \texttt{word+upos}, \texttt{word+xpos}, and \texttt{word+feats} vs \texttt{word}, using morphological tags predicted on perturbed and non-perturbed inputs.}
    \label{fig:gb_results}
\end{figure}

Figure \ref{fig:gb_results}.a shows the results for the graph-based parsers. Again, most morphological inputs contribute to degrade the performance faster than the baseline. 
In this case, no model beat the baseline when predicting tags on the perturbed inputs. The performance of \texttt{word+FEATS} and \texttt{word+UPOS} is similar (performing \texttt{word+UPOS} a bit better), and the \texttt{word+XPOS} models 
improve the performance the most. 

\begin{figure}
\centering
\begin{subfigure}{0.23\textwidth}
  \centering
  \includegraphics[width=\textwidth]{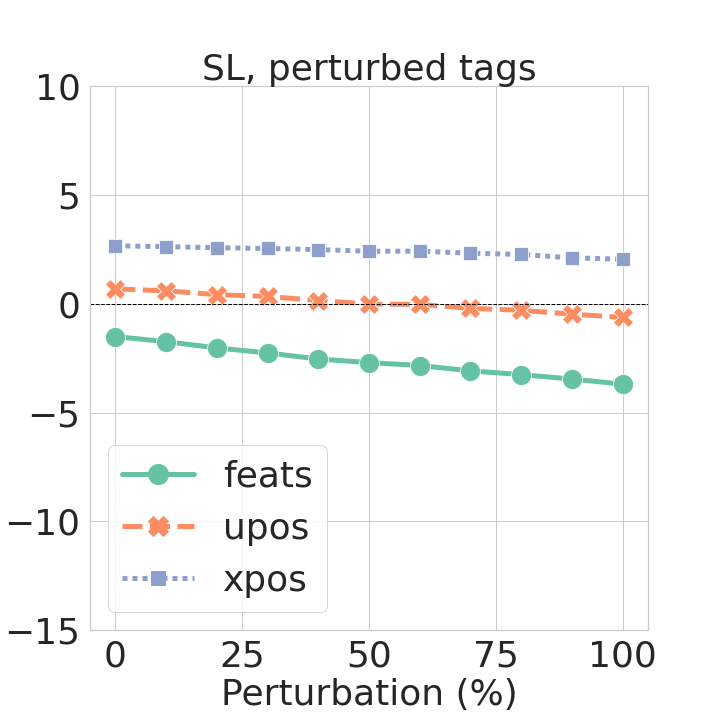}
  \caption{Perturbed}
\end{subfigure}
\hfill
\begin{subfigure}{0.23\textwidth}
  \centering
  \includegraphics[width=\textwidth]{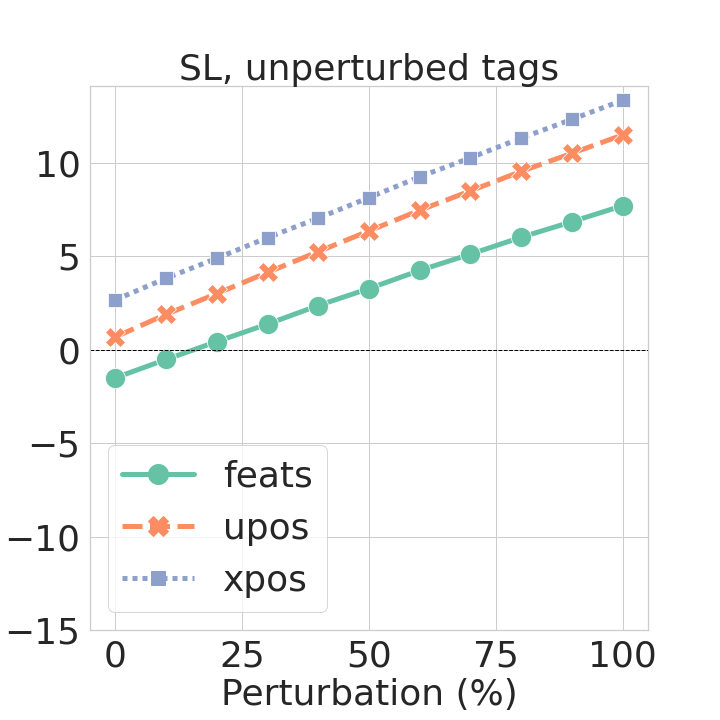}
  \caption{Non-perturbed}
\end{subfigure}
    \caption{Average $\Delta$LAS across all treebanks for the sequence-labeling models \texttt{word+upos}, \texttt{word+xpos}, and \texttt{word+feats} vs \texttt{word}, using morphological tags predicted on perturbed and non-perturbed inputs.}
    \label{fig:sl_results}
\end{figure}

Figure \ref{fig:sl_results}.a shows the results for the sequence labeling parsers:
differences between the baseline and the models utilizing morphological information exhibit minor changes ranging from 0\% to 100\% of perturbed words.
Also, the usefulness of the morphological information depends on the specific tags selected.
While \texttt{word+UPOS} obtains similar results to the baseline, 
\texttt{word+XPOS} scores around 2-3 points higher for the tested percentages of perturbations, and \texttt{word+FEATS} harms the performance in a range between 1 and 4 points.

The results show that feeding morphological tags  to both graph- and transition-based parsers has a negative 
impact to counteract such attacks, 
degrading their performance faster. 
On the contrary, the sequence labeling parsers, that rely on biLSTMs to make the predictions, can still benefit from them.
In addition, the different trends for the sequence labeling parser \emph{versus} the transition- and graph-based parsers, which additionally include a module to output trees (a pointer network and a biaffine attention, respectively), suggest that such modules are likely to be more effective against adversarial attacks than explicit morphological signals.

\subsubsection{Results using morphological tags predicted on non-perturbed inputs}
As mentioned above, we use this setup to estimate whether morphological tags could have a positive impact if they were extremely robust against lexical perturbations (see also Figures \ref{fig:tb_results}.b, \ref{fig:gb_results}.b and \ref{fig:sl_results}.b). In the case of the transition-based parser, we observe that morphological tags predicted on non-perturbed inputs help the parser
more as the inputs' perturbation grows, being \texttt{word+XPOS} the most helpful information, while \texttt{UPOS} and \texttt{FEATS} become useful only when sentences are perturbed over 20\% (but they also become more and more helpful). The graph-based parser also benefits from the use of more precise tags: \texttt{word+XPOS} models beat the baseline when the perturbation is over 30\%; and over 50\% for \texttt{word+UPOS} and \texttt{word+FEATS} setups.  Finally, for the sequence-labeling parser, morphological information from a robust tagger helps the model surpass the baseline for any percentage of perturbed words (except in the case of \texttt{word+FEATS}, when it only happens with perturbations over 20\%).

\subsubsection{Discussion on slightly perturbed inputs}
Unintended typos are commonly found among users.
For experiments with a small percentage of perturbed words ($<20\%$), transition-based parsers show improvement solely with the \texttt{word+XPOS} model, even when using non-robust taggers. Conversely, graph-based parsers do not benefit from morphological tags in this setup. Last, sequence labeling parsers benefit from incorporating \texttt{XPOS} and \texttt{UPOS} information, irrespective of the tagger's robustness, but not \texttt{FEATS}.

\subsubsection{Differences across morphological tags} Averaging across languages, the language-specific \texttt{XPOS} tags have a better (or less bad, for setup i) behavior. These tags are specific to each language. The coarse-grained \texttt{UPOS} tags have a common annotation schema and tagset. This eases annotation and understanding, but offer less valuable information. For \texttt{FEATS}, the annotation schema is common, but in this case they might be too sparse.

\section{Conclusion}
This paper explored the utility of morphological information to create stronger dependency parsers when these face adversarial attacks at character-level. Experiments over 14 diverse UD treebanks, with different percentages of perturbed inputs, show that using morphological signals help creating more robust sequence labeling parsers, but contribute to a faster degradation of the performance for transition- and graph-based parsers, in comparison to the corresponding word-only models. 

\section*{Acknowledgements}
This paper has received funding from 
grant SCANNER-UDC (PID2020-113230RB-C21) funded by MCIN/AEI/10.13039/501100011033, 
the European Research Council (ERC), which has supported this research under the European Union’s Horizon Europe research and innovation programme (SALSA, grant agreement No 101100615),
Xunta de Galicia (ED431C 2020/11), and Centro de Investigaci\'on de Galicia ``CITIC'', funded by Xunta de Galicia and the European Union (ERDF - Galicia 2014-2020 Program), by grant ED431G 2019/01.

\section*{Limitations}

\paragraph{Main limitation 1} The experiments of this paper are only done in 14 languages that use the Latin alphabet, and with a high share of Indo-European languages, with up to 4 Germanic languages. This is due to two reasons: (i) the scarcity of \texttt{XPOS} and \texttt{FEATS} annotations in treebanks from other language families, and (ii) the research team involved in this work did not have access to proficient speakers of languages that use other alphabets.
Hence, although we created a reasonable diverse sample of treebanks, this is not representative of all human languages.

\paragraph{Main limitation 2} Although we follow previous work to automatically generate perturbations at character-level, and these are inspired in psycholinguistic studies, they might not be coherent with the type of mistakes that a human will make. In this work, generating human errors  is not feasible due to the amount of languages involved, and the economic costs of such manual labour. Still, we think the proposed perturbations serve the main purpose: to study how morphological tags can help parsers when these face lexical errors, while the used method builds on top of most of previous work on adversarial attacks at character-level.

\bibliography{anthology,custom}
\bibliographystyle{acl_natbib}

\end{document}